\documentclass[journal]{IEEEtran}
\usepackage{ifpdf}

\ifCLASSOPTIONcompsoc
  \usepackage[nocompress]{cite}
\else
  \usepackage{cite}
\fi

\ifCLASSINFOpdf
  \usepackage[pdftex]{graphicx}
\else
  \usepackage[dvips]{graphicx}
\fi

\usepackage{amsmath}
\usepackage[linesnumbered,ruled]{algorithm2e}
\usepackage{array}
\usepackage{mdwmath}
\usepackage{mdwtab}
\usepackage{multirow}
\usepackage[caption=false,font=footnotesize,labelfont=sf,textfont=sf]{subfig}
\usepackage{enumerate}

\usepackage{url}

\begin{document}

\title{A Novel Training Protocol for Performance Predictors of Evolutionary Neural Architecture Search Algorithms}

\author{Yanan~Sun,~\IEEEmembership{Member,~IEEE,}
        ~Xian~Sun,
        ~Yuhan~Fang,
         and Gary~G.~Yen,~\IEEEmembership{Fellow,~IEEE} 

\thanks{Yanan Sun and Yuhan Fang are with the College of Computer Science, Sichuan University, Chengdu 610065, China (e-mails:ysun@scu.edu.cn; fangyuhan0719@gmail.com).}
\thanks{ Xian Sun is with Zhizhe Information Technology Services Company Limited, Chengdu 610000, China (e-mail:eric.x.sun@gmail.com).}
\thanks{Gary G. Yen is with the School of Electrical and Computer Engineering, Oklahoma State University, Stillwater, OK 74078 USA (e-mail:gyen@okstate.edu, corresponding author).}
}

\IEEEtitleabstractindextext{%
\begin{abstract}
Evolutionary Neural Architecture Search (ENAS) can automatically design the architectures of Deep Neural Networks (DNNs) using evolutionary computation algorithms. However, most ENAS algorithms require intensive computational resource, which is not necessarily available to the users interested. Performance predictors are a type of regression models which can assist to accomplish the search, while without exerting much computational resource. Despite various performance predictors have been designed, they employ the same training protocol to build the regression models: 1) sampling a set of DNNs with performance as the training dataset, 2) training the model with the mean square error criterion, and 3) predicting the performance of DNNs newly generated during the ENAS. In this paper, we point out that the three steps constituting the training protocol are not well though-out through intuitive and illustrative examples. Furthermore, we propose a new training protocol to address these issues, consisting of designing a pairwise ranking indicator to construct the training target, proposing to use the logistic regression to fit the training samples, and developing a differential method to building the training instances. To verify the effectiveness of the proposed training protocol, four widely used regression models in the field of machine learning have been chosen to perform the comparisons on two benchmark datasets. The experimental results of all the comparisons demonstrate that the proposed training protocol can significantly improve the performance prediction accuracy against the traditional training protocols.
\end{abstract}

}

\maketitle

\IEEEdisplaynontitleabstractindextext

\IEEEpeerreviewmaketitle

\section{Introduction}
\label{section_introduction}
  \IEEEPARstart{D}{eep} Neural Networks (DNNs) are becoming the dominant algorithm of machine learning~\cite{lecun2015deep}, largely owing to their superiority in solving challenging real-world applications~\cite{hinton2006reducing,krizhevsky2012imagenet}. Generally, the performance of DNNs relies on two deciding factors, the architectures of the DNNs and the weights associated with the architecture. The performance of a DNN in solving the corresponding problem can be promising, only when its architecture and the weights achieve the optimum combination simultaneously. Commonly, when the architecture of a DNN is determined, the optimal weights can be obtained through formulizing the loss as a continuous function, and then the exact optimization algorithms are employed for solving. In practice, the gradient-based optimization algorithms are the most popular one in addressing the loss function, although they cannot theoretically guarantee the global optimum~\cite{kearney1987optical}. On the other hand, obtaining the optimal architectures is not a trivial task because the architectures cannot be directly optimized as the weights do. In practice, most, if not all, prevalent state-of-the-art DNN architectures are manually designed based on extensive human expertise, including ResNet~\cite{he2016deep}, DenseNet~\cite{huang2017densely}, and among others.

Unfortunately, the expertise is often scarce in practice and generally owned by a very limited number of researchers and engineers who commonly work for industrial giants or major institutions. This is also the reason that state-of-the-art DNNs are often invented from such avenues. Given the promising performance of DNNs potentially leading to great economic benefits, many traditional industries are thirsting to employ DNNs to enhance their competitiveness in the market. However, they are also embarrassed in reality because 1) the architectures of DNNs are task-specified and the state-of-the-arts cannot be reused for their own problems; 2) they do not have sufficient expertise in designing the proper DNN architectures. Clearly, it will have a profound impact to make DNNs step further for a wider range of people without relevant expertise given an automatic mean in designing DNNs.

Neural Architecture Search (NAS) is the methodology that aims at automating the design of DNN architectures. Many encouraging results have shown that NAS is not only able to generate competitive architectures as human experts do, but also creates innovative architectures hardly known by human experts~\cite{elsken2018neural}. In principle, NAS is a complex optimization problem involving various challenges, e.g., complex constraints, discrete representations, bi-level structures, computationally expensive characteristics and multiple conflicting objectives~\cite{sun2019completely}. By solving NAS with the proper optimization algorithms, the DNN architecture design without human expertise can be realized. Existing NAS algorithms are mainly classified into three different categories based on their optimizers adopted, i.e., the Reinforcement Learning (RL)-based algorithms~\cite{baker2016designing,zoph2016neural}, the gradient-based algorithms~\cite{liu2018darts} and the Evolutionary Computation (EC)-based algorithms (ENAS)~\cite{xie2017genetic,real2017large,sun2019completely,liu2017hierarchical,suganuma2017genetic,sun2019evolving}. The RL-based algorithms often design the sub-components of a DNN, while the whole DNN must be combined from these sub-components through individual expertise, which is viewed as a kind of semi-automatic NAS algorithms~\cite{sun2019completely}. For the gradient-based algorithms, they often require a supernet, which is also manually designed with expertise in advance, and then the optimal DNN architecture is selected by choosing a path of the supernet. As the RL-based algorithms, the gradient-based algorithms are also viewed as the semi-automatic NAS algorithms. On the other hand, the ENAS algorithms are often fully automatic, and they can achieve the architecture design without any human intervention during searching for the optimal architectures of DNNs~\cite{sun2019completely,sun2019evolving}.

The EC algorithms belong to population-based heuristic computational paradigms~\cite{back1996evolutionary,banzhaf1998genetic,schmitt2001theory}. They have been widely used to solve discrete, constrained, bi-level, computationally expensive and conflicting, multi-objective problems, mainly because of their characteristics of gradient-free and insensitiveness to local minima~\cite{wang2015two_arch2,sun2018igd,twostage2019}. As evidenced from a recent survey of NAS~\cite{elsken2018neural}, most of the progress on NAS is largely made by EC. However, the major limitation of existing EC-based NAS algorithms lies in their enormous requirements on extensive computational resources. For example, on the benchmark dataset of CIFAR10~\cite{krizhevsky2009learning} which is very popular for verifying the performance of NAS algorithms, the Large-scale Evolution algorithm~\cite{real2017large} consumed 250 Graphic Processing Units (GPUs), while the Hierarchical Evolution algorithm~\cite{liu2017hierarchical} used 200 GPUs. In practice, such scale of the computational resource is not necessarily available to many interested users.

In principle, the heavy requirement of the computational resource for the ENAS algorithms is caused by the fitness evaluation of DNNs during the evolutionary search. Specifically, the fitness evaluation of a DNN is achieved by training the architecture on the target dataset via a training-from-scratch process, which is nevertheless time-consuming. For example, using one GPU to train a DNN on a common NAS benchmark, say CIFAR10, it often consumes hours to days depending on the scale of the DNN. Moreover, because the EC methods are population-based and there will be a number of individual DNNs to be evaluated during the search process of the ENAS. As a result, the ENAS becomes a prohibitively computationally expensive problem. Thus, the intensive computational resource is often required by ENAS algorithms to provide the solutions within an acceptable time. However, owning sufficient GPUs or renting remote cloud computing hours could be costly to most researchers. As a result, the researchers have proposed to use performance predictors, which are designed to predict the fitness values of DNNs by avoiding the time-consuming training process, to alleviate this computational issue.

The existing performance predictors can be generally classified into four different categories. They are the shallow training strategy-based performance predictors~\cite{sun2019evolving,wang2018evolving}, the learning curves-based performance predictors~\cite{swersky2014freeze,domhan2015speeding,klein2016learning}, the shared weights-based performance predictors~\cite{pham2018efficient} and the end-to-end performance predictors~\cite{istrate2018tapas,baker2017accelerating,deng2017peephole,sun2019surrogate}. Specifically, the ones based on the shallow training strategy take effect by training DNNs with a much smaller number of training samples or training epochs, and then obtain the performance by evaluating the DNNs directly on the validation dataset. These methods have been reported to have a poor generalization due to the insufficient training of DNNs~\cite{sun2019surrogate}. The performance predictors based on the learning curve work by building a regression model for each DNN based on the partial learning curve. Due to the requirement of smoothness for building the proper regression model, these methods perform poorly because the learning curves of DNNs are often non-smooth, which is caused by the scheduling of learning rates during the training of modern DNNs. The share weights-based performance predictors define a supernet in advance, and then the NAS is realized within the supernet. By training the supernet only once before running the NAS algorithm, the weights of the DNN are directly inherited from the supernet and are used to predict the performance. This sharing mechanism works with the assumption that the weights of a DNN are dependent with each other. However, it is easy to prove that the dependence exists only for two-layer NNs. The end-to-end performance predictors work by collecting a group of DNNs with their corresponding performance, and then build a regression model to map the DNN architectures and the performance values. When a new DNN has been generated during the evolutionary search, the regression model directly predicts its performance. The end-to-end performance predictors are principally characterized without the limitations suffered by the peer competitors. Recent progresses~\cite{istrate2018tapas,baker2017accelerating,deng2017peephole,sun2019surrogate} on performance predictors are all falling into this category.

Although different end-to-end performance predictors have been developed during the past years, they follow similar training protocols. Firstly, a set of DNNs is sampled and then trained to obtain their respective performance values. Then, a linear regression model is built based on the DNN architectures and their performance values using the criterion of Mean Square Error (MSE).  Finally, the built regression model is used to predict the performance of newly generated DNNs. Through replacing the computational expensive training of DNNs by this performance predictor, the requirement on the massive computational resource of ENAS algorithms can be greatly alleviated. However, this traditional training protocol bears multiple limitations. First of all, because the MSE concerns only the absolute values between the target and the predicted output, the resulted regression model may mislead the ENAS. In addition, it is hard to build an exact linear regression model for the mapping because the training samples have no particular order resulting the same training data. Furthermore, directly using the raw forms of the training samples cannot provide satisfactory prediction result (we will detail these three limitations in Subsection~\ref{sec2_2}).  

In this paper, we propose a new training protocol for performance predictors in ENAS, to address all the limitations aforementioned. The contributions of the proposed work are summarized below:  
\begin{itemize}
	\item A Pairwise Ranking Indicator (PRI) is proposed in this paper, to address the limitation caused by the MSE used by the traditional training protocols. Specifically, in the proposed PRI, the ranking information between any two training samples is used to train the regression model. The proposed PRI is consistent with the selection principles in ENAS algorithms, and its result can directly reflect the ordering of candidates without any further comparison. As a result, the proposed PRI will effectively guide the search of ENAS.
	
	\item We propose to use the logistic regression to replace the linear regression in building the performance predictors. By doing so, the proposed PRI can be reasonably built upon the logistic regression model. Furthermore, because the logistic regression model does not concern the order of the training samples, the corresponding problem caused by the linear regression model upon the traditional training protocol will be solved accordingly.
	
	\item A differential method has been developed to adapt to the best practice of ENAS algorithms. Specifically, we use the difference between any pair data as the training instances to train the performance predictor, so that the trained model still works well for the situation that the unseen data did not lie in the range of the training data. In addition, the proposed method can create a couple of balanced training samples for the given pair data. Therefore, the data imbalanced problems, which leads to the great challenge to most machine learning tasks, is no longer in existence.
\end{itemize}

The remainder of this paper is organized as below. The background of the traditional training protocol is discussed in Section~\ref{sec2}. Immediately after, the details of the proposed work are documented in Section~\ref{sec3}. This is followed by the experimental setup and the experiment results, which are provided in Sections~\ref{sec4} and~\ref{sec5}, respectively. Finally, the conclusion and future work are outlined in Section~\ref{sec6}.

\section{Background}
\label{sec2}
To help the readers easily understand the work proposed in this paper, we will first briefly introduce the ENAS in Subsection~\ref{sec2_1}. Particularly, the fitness evaluation and the selection operation of ENAS algorithms will be described in detail, to justify the necessity of the proposed work. Based on which, the limitations of the traditional training protocol, which are widely used to train the performance predictors, are explained in Subsection~\ref{sec2_2}.

\subsection{Evolutionary Neural Architecture Search (ENAS)}
\label{sec2_1}
	ENAS refers to using Evolutionary Computation (EC) methods to realize the NAS. Generally, an ENAS algorithm constitutes of the man steps shown below:

\begin{enumerate}[Step 1:]
	\item Initialize a population of individuals representing different DNN architectures within the predefined search space;\label{enas_step1}
\item Employ a massive amount of computation resource to evaluate the fitness of each DNN architecture;\label{enas_step2}
\item Select the parent solutions from the population based on the fitness values;\label{enas_step3}
\item Generate new offspring DNNs with the parent solutions using genetic operators;\label{enas_step4}
\item Go to Step~\ref{enas_step2} if the evolutionary generation does not exceed the predefined maximal generation number; otherwise, go to Step~\ref{enas_setp6};
\item Terminate the evolutionary process and output the DNN having the best fitness value.~\label{enas_setp6}
\end{enumerate}

As can be seen above, the ENAS algorithm follows the standard flow chart of an EC method. Specifically, Step~\ref{enas_step2} shows the fitness evaluation of the ENAS algorithms, where a population of DNN architectures is trained. This step is generally performed on a cluster of GPU computers because GPUs could significantly speed up the training of DNNs compared to the CPU-based computation. In addition, the fitness values may refer to different meanings that are decided by the problems on which the ENAS algorithm targets. For example, if the ENAS is for image classification tasks, the fitness value can refer to the classification accuracy of the DNNs on the image dataset. The proposed training protocol is based on performance predictors, and the performance predictors are for the fitness evaluation and the selection of the corresponding ENAS algorithms. In the following, we will discuss them in detail, to help understand how the training protocol works, including both the traditional training protocol and the proposed training protocol.

\subsubsection{Fitness Evaluation of ENAS Algorithms}
\begin{algorithm}
	\label{alg_enas_fitness_evaluation}
	\caption{Fitness Evaluation of ENAS Algorithms}
	\KwIn{The population $\mathcal{Q}$ for the fitness evaluation, the training dataset $\mathcal{D}_{train}$, the fitness evaluation dataset $\mathcal{D}_{fitness}$, the bacth number $batchs$, the epoch $epochs$.}
	\KwOut{The population $\mathcal{Q}$ after the fitness evaluation.}
	\ForEach{individual $q$ in $\mathcal{Q}$}
	{\label{alg_enas_fitness_for1}
		$c\leftarrow$ Decode $q$ to the corresponding CNN\;
		\For{i = 0 to $epoch$}
			{\label{alg_enas_fitness_for2}
				\For{j = 0 to $batch$}
				{\label{alg_enas_fitness_for3}
					Train $q$ on the $j$-th batch data of $\mathcal{D}_{train}$\;
				}
			}
		$v\leftarrow$ Evaluate $c$ on $\mathcal{D}_{fitness}$\; 
		Set $v$ as the fitness of $q$\;
	}
	
	\textbf{Return} $\mathcal{Q}$.
\end{algorithm}

Algorithm~\ref{alg_enas_fitness_evaluation} shows the framework of the fitness evaluation of ENAS algorithms. Generally, the fitness evaluation of EC methods refers to calculating the fitness value of each individual in the population. In the context of ENAS algorithms, the evaluation means to obtain the performance of the DNNs by individually training them on the corresponding dataset. As can be seen from Algorithm~\ref{alg_enas_fitness_evaluation}, there are three for-loop nests to accomplish the fitness evaluation of the ENAS algorithms. The first is to locate each individual in the population (Line~\ref{alg_enas_fitness_for1}), the second is the training of the individual for each epoch (Line~\ref{alg_enas_fitness_for2}), and the third is the training of the individual on each batch data (Line~\ref{alg_enas_fitness_for3}). With these three nests, there will be millions of training phases in the ENAS algorithm, which is highly expensive. For example, for the commonly used CIFAR10 benchmark dataset, the training epochs are often specified as $500$, the batch number is 391 (the CIFAR10 dataset has 50,000 samples, and the batch size is often set as 128, thus, $\lceil 50,000/128\rceil=391$, where $\lceil  \cdot \rceil$ is a ceiling operator). To this end, each individual in the population will be trained with $2\times10^5$ times approximately. Commonly, training a common CNN on CIFAR10 per epoch consumes about two minutes given the use of one GPU card. Thus, the CNN will take 17 hours to finish its training on the whole dataset. In a traditional ENAS algorithm, the number of $1,000$ individuals (this is calculated by multiplying the generation number and the population size predefined in the ENAS algorithm) is a very commonly used configuration for the individuals to be trained in the ENAS algorithm. As a result, the ENAS algorithm will consume 700 days approximately in this situation. In the case of using a larger size of population and generation for a larger-scale dataset, this consumption will become prohabitively challenging. This is also the reason that most ENAS algorithms are performed on a computation platform equipped with a large number of GPU cards. For example, if we employ 20 GPU cards for the fitness evaluation, the above ENAS algorithm will only need 35 days instead, which will be acceptable compared to the 700 days using one GPU card. However, due to the high cost of GPU cards, the intensive computational resource is not necessarily available to each of the users related. Performance predictors can faithfully predict the performance of the DNNs without the inner two for-loop nests, which are also the most time-consuming phases in the ENAS algorithms. This is the motivation of developing proven performance predictors.

\subsubsection{Selection Operation of ENAS Algorithms}
The selection operation performs the biological principle of EC methods, i.e., the principle of ``survival of the fittest''. Because the ENAS algorithms are designed based on EC methods, they also need to perform the selection operation. Generally, in an ENAS algorithm, there are two different phases on which the selection operation take effects. The first is about the genetic selection, aiming at selecting parent solutions from the current population for generating new offspring, while the other is the environmental selection to select individuals from the combined population of both the offspring and the parents, to form the parent population of the next generation.

There are two frequently used selection operators based on selection behaviour. The first is the roulette wheel selection~\cite{de1975analysis}, which selects the solutions based on the probabilities in proportion to their fitness values, i.e., a solution having higher fitness value will be selected with a higher probability. The other is the tournament selection~\cite{miller1995genetic} that selects the solutions based on the ordering of the fitness values of the candidates, i.e., the one having the highest fitness value among the candidates will be favored. In principle, the roulette wheel selection is more suitable for the individuals having discrete fitness values, which will benefit to the calculation of the probabilities for the selection. Because the fitness values in ENAS algorithms are often continuous values (such as the classification accuracy), most, if not all, of the existing ENAS algorithms employ the tournament selection operator.

Based on the traditional training protocol, the prediction results of the performance predictors are the fitness values of the individuals, and then the selection is performed by comparing the ordering of the fitness values of the candidates. As have discussed above that the selection is directly based on the ordering of the individuals, the commonly used selection operator in ENAS algorithms does not necessarily require the fitness values. The traditional training protocol employs the MSE to train the performance predictors, which is not the real scenario of using performance predictor. In addition, using the MSE as the criterion of training the performance predictors also suffer from several limitations in practice (it will be discussed in Subsection~\ref{sec2_2}), which cannot guarantee the promising prediction results. Commonly, the binary tournament selection is often used, i.e., the number of candidates is set two. When the best CNN architecture is selected from the final population, i.e., the selection of elitism, we directly view the whole population as the candidates.

\subsection{End-to-End Performance Predictors}
\label{sec2_2}
The existing end-to-end performance predictors are almost all linear regression models, which are trained on a set of DNN architectures with their performance measures through minimizing the MSE of the predicted values and the performance values trained in advance. Note that the DNN architectures should be sampled by the same ENAS algorithm on which the performance predictor is going to work. In addition, the performance of the DNNs should be obtained like those shown in Step~\ref{enas_step2} of Subsection~\ref{sec2_1}. In Section~\ref{section_introduction}, the three major limitations of the end-to-end performance predictors have been summarized. In this subsection, we will provide the justification in detail.

Firstly, the regression model optimized by minimizing the MSE may mislead the ENAS algorithms because MES concerns only the absolute values. Particularly, the fitness of individuals in EC is mainly used for the selection, i.e., the genetic selection and the environment selection, to select the individuals having better fitness values. Specifically, the selection is based on the criterion that which one has the better fitness value. Because of the absolute value resulted by MSE, the predicted fitness value cannot be compared reasonably. For example, given two DNNs represented by $d_i$ and $d_k$ whose ground-truth fitness values are $0.4$ and $0.5$, respectively. Supposing their predicted fitness values are $0.45$ and $0.44$, respectively. These predicted values are reasonable, although they are not consistent with their true values when used for the order comparison. This is because the predicted values satisfy the MSE training criteria, i.e., the MSE of $0.4$ and $0.45$ is less than that of $0.44$ and $0.5$. To this end, with the traditional training protocol, $d_i$ will be selected because it has a smaller MSE. However, since $d_k$ has a better fitness value, it should be chosen by the selection operation. Clearly, this will mislead the selection operation of the ENAS algorithms, which in turn inevitably deteriorates the performance in searching for the optimal CNN architectures.

\begin{figure}[!htp]
	\centering
	\subfloat[]{\includegraphics[width=0.95\columnwidth]{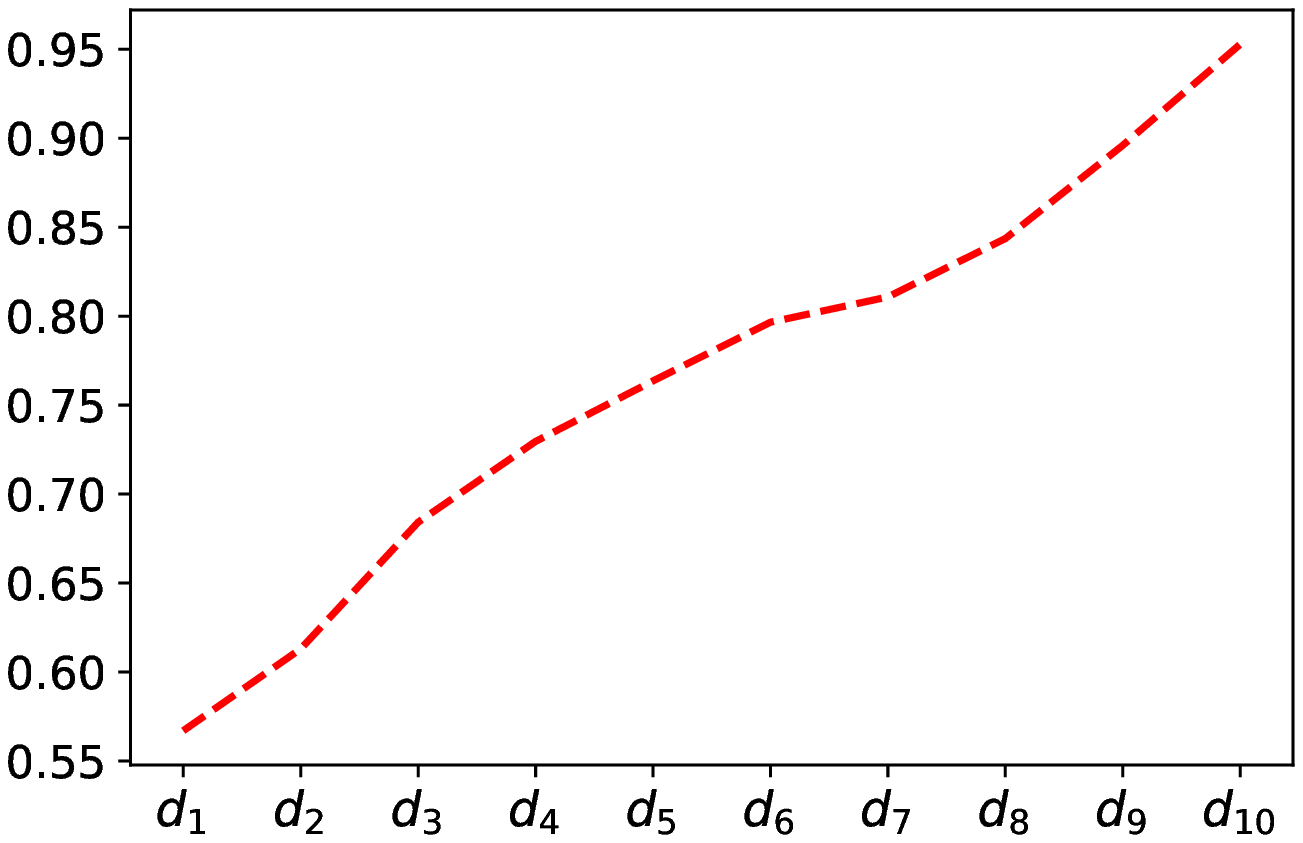}\label{fig_normal_example}}
	\hfil
	\subfloat[]{\includegraphics[width=0.95\columnwidth]{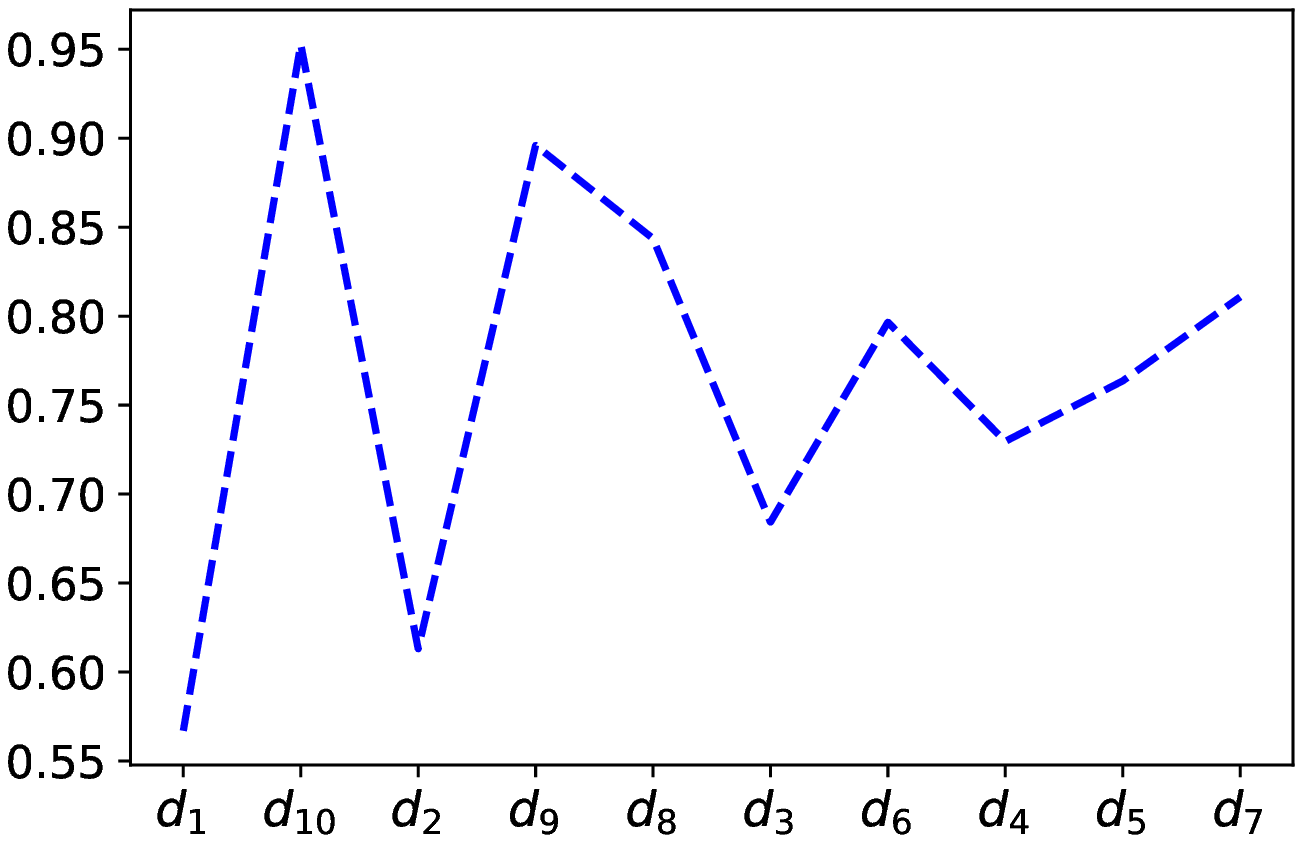}\label{fig_compared_example}}
	\caption{An example to illustrate the significantly different curves resulted by the sample samples but with different orders. In both figures, the vertical axis denotes the names of the $10$ DNNs, and the horizontal axis demote their corresponding performance.}\label{fig_regression_example}
\end{figure}

Secondly, the regression model cannot be accurately built because the training samples have no particular order when using the traditional training protocol. Specifically, the regression models are almost linear to fit the performance curves of the training samples. Generally, the performance of the linear regression models can be promising only when the curves are smooth. However, the shape of the curves is decided by the order of samples in training the regression model. The training samples have no particular order, and different users may apply different orders of the training samples for training the regression models. This will result in different regression models for the same training samples, and provide different prediction results for the same DNN. 

To make readers intuitively better understand this limitation, an illustrative example is provided in Fig.~\ref{fig_regression_example}, showing $10$ DNNs (denoted by $d_1$ to $d_{10}$) associated with their respective performance value. Specifically, Figs.~\ref{fig_normal_example} and~\ref{fig_compared_example} show the curves of the performance from the same DNNs but with two different orders. As can be seen from both figures, the curve shown in Fig.~\ref{fig_normal_example} is much smoother than that shown in Fig.~\ref{fig_compared_example}, and the linear regression model for the curve shown in Fig.~\ref{fig_normal_example} will be much easily built than that shown in Fig.~\ref{fig_normal_example}. However, both figures show the curves of the same data. Please note that the order can be manually assigned to address this problem, for example, ordering these training samples based on their performance in an increasing order. However, this assignment has no theoretical proof. Moreover, this assignment cannot provide satisfying results in practice as well~\cite{sun2019surrogate}.

Thirdly, for obtaining a high prediction accuracy, the end-to-end performance predictors often sample a large number of DNNs as the training data, where the sampling is performed with an independent running of the ENAS algorithm. However, this way is not consistent with the real application scenario how the performance predictors work. Specifically, in an ENAS algorithm, the individuals in the first evolutionary process are often with lower fitness. As the evolutionary process proceeds, more and more individuals are with higher fitness. When the training data are obtained with an independent run, the collected data may have contained the best individuals, and the best CNN architecture has been found. Thus, there is no need to perform the performance predictor. For example, the Peephole method~\cite{deng2017peephole} sampled $8,000$ DNN architectures for training its performance predictor. However, in a typical ENAS algorithm, there will be about 500 DNN architectures evaluated. Clearly, it is not wise sampling a larger number of DNNs than an ENAS algorithm normally dose. In principle, the sampled DNNs should be with the smaller fitness values, and the individuals whose fitness to be predicted should be the one with higher fitness values, and the ranges of their fitness values are without overlap. However, this will also give rise to another challenge to the linear regression model that the prediction is based on the interpolation. If the fitness values of the trained data do not cover those of the individuals to be predicted, the interpolation results will be poor. 

\begin{figure}[!htp]
	\centering
	\includegraphics[width=0.95\columnwidth]{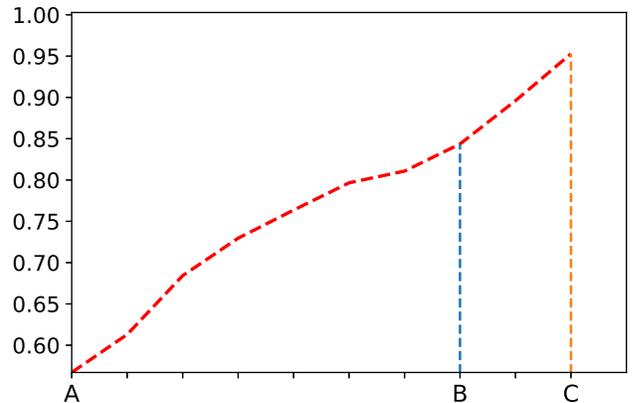}
	\caption{An example to illustrate the third limitation of the currently popular training protocol in directly using the sampled DNNs as the training data of performance predictors.}\label{fig_range_example}
\end{figure}

An example shown in Fig.~\ref{fig_range_example} is used to illustrate this situation. Specifically, Fig.~\ref{fig_range_example} shows the performance values of the sampled DNNs, where the curve between $A$ and $C$ can be segmented into two parts: the curve between $A$ and $B$, and the curve between $B$ and $C$. If the performance predictor is built based on the curve between $A$ and $B$, the predicted result is reasonable only when the ground truth performance of the newly generated DNN lies in the range between $A$ and $B$. This is because the prediction is based on the interpolation. In practice, the newly generated DNN should be within the range between $B$ and $C$, resulting in the predicted value that cannot go beyond the curve between $A$ and $C$. Clearly, the prediction result will not be accurate. The main reason is caused by directly using the raw data of the sampled architectures to train the performance predictors. 

\section{The Proposed Training Protocol}
\label{sec3}
In this section, we will document the details of the proposed training protocol. Specifically, we will first provide the overall framework of the proposed training protocol in Subsection~\ref{sec3_1}, and then introduce the main steps in Subsections~\ref{sec3_2} to \ref{sec3_4}.

\subsection{Framework Overview}
\label{sec3_1}
\begin{algorithm}
	\label{alg_framework}
	\caption{The Proposed Training Protocol}
	\KwIn{A set of sampled DNN architectures and their respective performance values.}
	\KwOut{The trained performance predictor.}
	$\mathcal{X}\leftarrow$ Construct the training instances based on provided the DNN architectures using the proposed differential method\;\label{alg_frame_step1}
	$\mathcal{Y}\leftarrow$ Construct the training labels based on the provided DNN performance using the proposed pairwise ranking indicator\;\label{alg_frame_step2}
	$\mathcal{D}=\{\mathcal{X}, \mathcal{Y}\}\leftarrow$ Build the training data\;\label{alg_frame_step3}
	$\mathcal{R}\leftarrow$ Initialize a logistic regression model\;\label{alg_frame_step4}
	Train $\mathcal{R}$ using $\mathcal{D}$\;\label{alg_frame_step5}
	\textbf{Return} $\mathcal{R}$.\label{alg_frame_step6}
\end{algorithm}

Algorithm~\ref{alg_framework} shows the framework of the proposed training protocol, where the sampled DNN architectures and their training performance values are given in advance, and then the proposed protocol takes effect. At last, the trained performance predictor is obtained for use. Specifically, the proposed training protocol is composed of two parts, i.e., the construction of the training data and the training of the performance predictors, which are shown in Lines~\ref{alg_frame_step1} to \ref{alg_frame_step2} and Lines~\ref{alg_frame_step3} to \ref{alg_frame_step5} of Algorithm~\ref{alg_framework}, respectively. Particularly, in the first part, the training instances are first constructed by the proposed differential method based on the provided DNN architectures (Line~\ref{alg_frame_step1}), and then the training labels of the corresponding training instances are constructed by the proposed pairwise ranking indicator (Line~\ref{alg_frame_step2}). In the second part, the training data are build based on the instances and the corresponding labels first (Line~\ref{alg_frame_step3}), and then a logistic regression model is initialized as the performance predictor (Line~\ref{alg_frame_step4}). Through training the regression model with the built training data (Line~\ref{alg_frame_step5}), the performance predictor is obtained for the corresponding ENAS algorithm (Line~\ref{alg_frame_step6}).

The performance predictors are ENAS-specific, i.e., different ENAS algorithms generally have different performance predictors. This can be understood like that we often employ DNN models for building image classification models. However, for different types of image datasets, we need to train the DNN model on the corresponding dataset, although all the models are with the same DNN architectures. As a result, for the proposed training protocol, the given DNN architectures should follow the potential architectures generated by the corresponding ENAS algorithm, and their performance should also be obtained by following the same training routine as the ENAS algorithm has. Principally, for doing so is to ensure to follow the same data distribution between the training phase and the test phase of machine learning algorithms. 

\subsection{Constructing Training Instances}
\label{sec3_2}
The performance predictors are supervised learning models. In order to train such kind of supervised models, the training dataset is required. Particularly, the are a number of training samples in the training dataset, and each sample is composed of the instance and its corresponding label. This subsection will discuss how the instances are constructed in the proposed training protocol to effectively and efficiently train the performance predictors. 

\begin{algorithm}
	\label{alg_build_instance}
	\caption{The Proposed Differential Method}
	\KwIn{The sampled DNN architectures $\mathcal{A}=\{a_1,a_2,\cdots,a_n\}$.}
	\KwOut{The constructed instances $\mathcal{X}$.}
	$\mathcal{V}=\{v_1,v_2,\cdots,v_n\}\leftarrow$Vectorize each architecture in $\mathcal{A}=\{a_1,a_2,\cdots,a_n\}$\;\label{alg2_step1}
	$\mathcal{X}\leftarrow$ $\emptyset$\;\label{alg2_step2}
	\For{$i=1$ to $n-1$}
	{\label{alg2_step3_start}
		\For{$j=i+1$ to $n$}
		{
			$v_{i,j}\leftarrow$ $v_i-v_j$\;\label{alg2_step3_1}
			$v_{j,i}\leftarrow$ $v_j-v_i$\;\label{alg2_step3_2}
			$\mathcal{X}\leftarrow$ $\mathcal{X}\cup v_{i,j}\cup v_{j,i}$\;\label{alg2_step5}
		}
	}\label{alg2_step3_end}
   \textbf{Return} $\mathcal{X}$.

\end{algorithm}

The instance construction is achieved by the proposed differential method, and the details are shown in Algorithm~\ref{alg_build_instance}. Firstly, each architecture of the sampled DNNs is vectorized into the form which can be accepted by a computer program (Line~\ref{alg2_step1}), and then the $\mathcal{X}$ is initialized as an empty set (Line~\ref{alg2_step2}), to carry all the instances of which the construction details are shown in Lines~\ref{alg2_step3_start} to \ref{alg2_step3_end}. In the phase of instance construction, any two sampled architectures are collected first, and then their difference is employed as the instance (Lines~\ref{alg2_step3_1}-\ref{alg2_step3_2}). Note that the difference will result in two instances based on the order of the two collected architectures. Because the differences of the sampled architectures are used as the instances, this is also the reason why this method is named as ``differential method''. 

In principle, any vectorization method can be used to vectorize the architectures of the DNNs sampled. The reason of vectorization is that the sampled architectures are in the form of natural language-based representation, such as a paragraph of architecture description, or a segment of codes represented by a particular programming language, such as Tensorflow~\cite{abadi2016tensorflow} or PyTorch~\cite{paszke2019pytorch}. However, such representations of DNN architectures cannot be directly used by computer systems. The goal of the vectorization is to transform the architectures represented with languages or particular implementations to the one that the computer programs can recognize. The vectorization results are generally in the form of a group of numbers. This is like the natural language processing technique where the natural language must be transformed into vectors, such as the one-hot encoding method, so that the computer can process. To the best of our knowledge, there is no solution at hand that can vectorize DNN architectures in a unified way. In practice, the vectorization method is designed by the developers of the corresponding ENAS algorithms with which the trained performance predictors will work. For example, the Peephole algorithm has proposed the 4-tuple method to vectorize the DNN architectures which are realized by their proposed NAS algorithm, while the E2EPP method proposed a special vectorization method for the NAS algorithm that directly searches the DNN architectures based on blocks. Note that, the difference operator, i.e., the minus symbols (shown in Lines~\ref{alg2_step3_1} and \ref{alg2_step3_2}), operate on element-wise. For example, if the architecture is vectorized to a tuple with $k$ elements, the resulted difference should also be a tuple having $k$ elements. Clearly, $\mathcal{X}$ is with the size of $2n(n-1)$ given the $n$ sampled DNNs. In addition, each couple of architectures construct two instances, which will be justified in Subsection~\ref{sec3_3}.

As can be seen from Algorithm~\ref{alg_build_instance}, the instances used to train the performance predictors are the differences between the vectorization of any two sampled DNN architectures, instead of the vectorization result. As have discussed in Subsection~\ref{sec2_2}, the performance of any machine learning algorithms is promising only when the training data and the test data have the same data distribution. In the real application scenario of performance predictors for ENAS algorithms, the sampled architectures for the training data should be from the early generation, and commonly with lower performance. However, for the CNN architectures of which the performance values need to be predicted by the predictors, they are often generated in the later generations and are with higher performance, i.e., the CNNs used for training the performance predictors and the CNNs using the performance predictor are with different data distribution. Therefore, directly using the sampled architectures by the traditional training protocol to train the performance predictors is not reasonable. In the proposed training protocol, we use the difference between any two sampled architectures to train the performance predictors. Clearly, the difference between the sampled architectures and those architectures to use the performance predictors are probably with the same data distribution. We still use the example shown in Fig.~\ref{fig_range_example} to further clarify this point. Specifically, the difference of the performance between the sampled architectures (i.e., the points between $A$ and $B$) are from $0.14$ to $0.25$, while those of the architectures which will use the performance predictors (i.e., the points between $B$ and $C$) vary from $0.05$ to $0.11$. Clearly, the data distribution of the later one is within that of the former one when using the difference as the training instances. However, if using the way of the traditional training protocol, i.e., directly using the performance of the architectures as the training data, the performance range of points $A$ and $B$ (from $0.55$ to $0.83$) has no overlap with that of points $B$ and $C$ (from $0.83$ to $0.95$). This will clearly give rise to the poor performance of the performance predictor resulted.

\subsection{Constructing Training Labels}
\label{sec3_3}
\begin{algorithm}
	\label{alg_build_label}
	\caption{The Proposed Pairwise Ranking Indicator (PRI)}
	\KwIn{The perfromance $\mathcal{P}=\{p_1,p_2,\cdots,p_n\}$ for the sampled DNN architectures $\mathcal{A}=\{a_1,a_2,\cdots,a_n\}$.}
	\KwOut{The constructed labels $\mathcal{Y}$.}
	$\mathcal{Y}\leftarrow \emptyset$\;
	\For{$i=1$ to $n-1$}
	{\label{alg3_step3_start}
		\For{$j=i+1$ to $n$}
		{
			\If{$p_i - p_j \geq 0$}
			{
				$\mathcal{Y}\leftarrow$$\mathcal{Y}\cup \{1, 0\}$ \label{alg3_step5}
			}
			\If{$p_i - p_j < 0$}
			{
				$\mathcal{Y}\leftarrow$$\mathcal{Y}\cup \{0, 1\}$\label{alg3_step6}
			}
		}
	}\label{alg3_step3_end}
	\textbf{Return} $\mathcal{Y}$.
\end{algorithm}

The training labels are part of the training dataset for supervised machine learning algorithms, playing the role of the supervisor during the training phase. In the proposed training protocol, the labels are constructed by the proposed PRI, and its details are shown in Algorithm~\ref{alg_build_label}. Given the trained performance of the sampled DNN architectures, denoted by $\mathcal{P}$, an empty set, say $\mathcal{Y}$, is initialized first, and then every two values in $\mathcal{P}$ are paired. Supposing the performance values of the paired data are $p_i$ and $p_j$, the labels of $1$ and $0$ will be generated if $p_i$ is not less than $p_j$; otherwise, the labels of $0$ and $1$ are constructed. There are a couple of motivations behind this design.

Firstly, the proposed PRI is used as the supervising signals to effectively train the performance predictors. As a result, the predicted result is also about the PRI. This is different from that in the traditional training protocol, where the prediction results are the performance of the architectures. We argue that using the PRI during the training and the prediction phases is more reasonable than using the MSE of the traditional training protocols. Specifically, in the ENAS algorithms, the performance of the architectures is used for the selection, to select the one having a larger fitness value. When the traditional training protocol is used, the selection is performed through two steps: 1) predicting the performance of the candidates and 2) selecting the one having the higher performance. If we use the PRI, we do need to perform the two steps, instead of directly picking up the one having a higher performance by predicting their order, i.e., the use of PRI is a direct way to coincide the goal of the predicted performance. For example, if the ENAS chooses two candidates from the DNN architectures, which are denoted by $a_i$ and $a_j$. In the traditional situation where the performance is predicted by the performance predictors, we firstly predict their performance, say $p_i$ and $p_j$, respectively. In the ENAS algorithms, $p_i$ and $p_j$ are used to compare $a_i$ or $a_j$ who has the larger fitness value. If the proposed PRI is used, the predicted result is also about the PRI, i.e., directly giving the result that $a_i$ or $a_j$ has a better fitness value. As shown in Algorithms~\ref{alg_build_instance} and \ref{alg_build_label}, where the construction of the instances and the labels for collectively building the training samples, the PRI can be formalized as a binary classifier, where if the performance of one is not less than another one, we provide their difference as a positive value, and vice visa.

Secondly, as shown in Lines~\ref{alg3_step5} and \ref{alg3_step6} of Algorithm~\ref{alg_build_label}, a positive label and a negative label are generated for the two same architectures, which also corresponds to the constructions of the instances shown in Line~\ref{alg2_step5} of Algorithm~\ref{alg_build_instance}. Particularly, for the sampled architectures $a_i$ and $a_j$ whose vectorization representations are $v_1$ and $v_2$, respectively, we construct two instances regarding the differences calculated by $v_i-v_j$ and $v_j-v_i$, respectively, and then provide a positive label for the difference not less than zero, otherwise, the negative label is given. The reason for doing so is to avoid the data imbalance problem which widely exists and often gives rise to most machine learning tasks. In the proposed training protocol, we avoid this problem by constructing the samples of training performance predictors. Thus, the performance of the accomplished performance predictors can be principally guaranteed.

\subsection{Training Performance Predictors}
\label{sec3_4}

Training the performance predictors using the proposed training protocol is straightforward. It adopts similar training processes as those for other machine learning algorithms. Evidenced by the details of constructing the training instances and labels, the training samples are built for the binary classification tasks. As a result, the performance predictor should be about logistic regression models. The proposed training protocol is designed to work with any logistic regression model, and the researchers can decide it with their preferences For example, some researchers prefer using neural network-based models, while other likes the ensemble models. Because the training phase is straightforward, the details will not be provided. Here, we provide a guideline to choose the regression models for the training of performance predictors within the context of the proposed training protocol.

As the fitness evaluation of the ENAS algorithms, the sampling of DNN architectures for training the performance predictors is computationally expensive. As a result, the number of collected samples is limited, resulting in inadequate training samples for effectively training the regression model. In this case, we suggest the use of ensemble models. Firstly, the ensemble models work well in practice when the training data is limited. Though the neural network-based learning algorithms are very popular recently, their performance highly relies on a large number of training samples. Secondly, compared to the neural network-based learning algorithms, the ensemble models often have fewer hyper-parameters, thus it would be more computationally affordable for the parameter tuning. Note that the proposed training protocol considers the difference between any two samples, which could only predict the ordering of both. However, for the elitism selection and the tournament selection which are not binary (e.g., tournament selection with the candidate number more than two), they can still be achieved by performing multiple times of the performance predictors trained with the proposed training protocol.

\section{Experimental Setup}
\label{sec4}
To verify the performance of the proposed training protocol, a series of well-designed experiments are conducted. In the experiments, the traditional training protocol will be used as the baseline for the comparisons against the proposed training protocol. In the following, we will firstly introduce the dataset used for the experiments in Subsection~\ref{sec4_1}. Then, the regression models used to accomplish the training of the performance predictors are detailed in Subsection~\ref{sec4_2}. This is following by the parameter settings involving the experiments in Subsection~\ref{sec4_3}, and finally, the performance metrics are provided in Subsection~\ref{sec4_4}.

\subsection{Benchmark Datasets}
\label{sec4_1}
As have discussed above, the work proposed in this paper is a new training protocol, aiming at training effective and efficient performance predictors for the ENAS algorithms. Because almost all performance predictors are regression models in nature, the proposed training protocol will particularly train regression models on a set of DNN architectures sampled by ENAS algorithms, and then the trained model is used to predict the performance of DNN architectures newly generated in the ENAS algorithm. As a result, a set of sampled DNN architectures should be prepared in advance to serve as the training data in this experiment.

In principle, any DNN architectures can be adopted for the training data, so long as the training data is sampled from the ENAS algorithm, which will use the corresponding performance predictor. In this experiment, we use the architectures sampled from the AE-CNN algorithm~\cite{sun2019completely} given the considerations as following. Firstly, the AE-CNN algorithm is a state-of-the-art ENAS. Using its generated CNN architectures will have the representative meaning. Secondly, the architectures have been further used to design the E2EPP algorithm~\cite{sun2019surrogate}, which is a state-of-the-art performance predictor for ENAS algorithms. Thirdly, the E2EPP method has proposed an efficiency vectorization method for the CNN architectures. This could help to focus on extensively demonstrating the performance of the training protocol, without exerting any concern on the design of vectorization method for the CNN architectures.

\begin{figure*}[!htp]
	\centering
	\subfloat[CIFAR10-CNN]{\includegraphics[width=0.9\columnwidth]{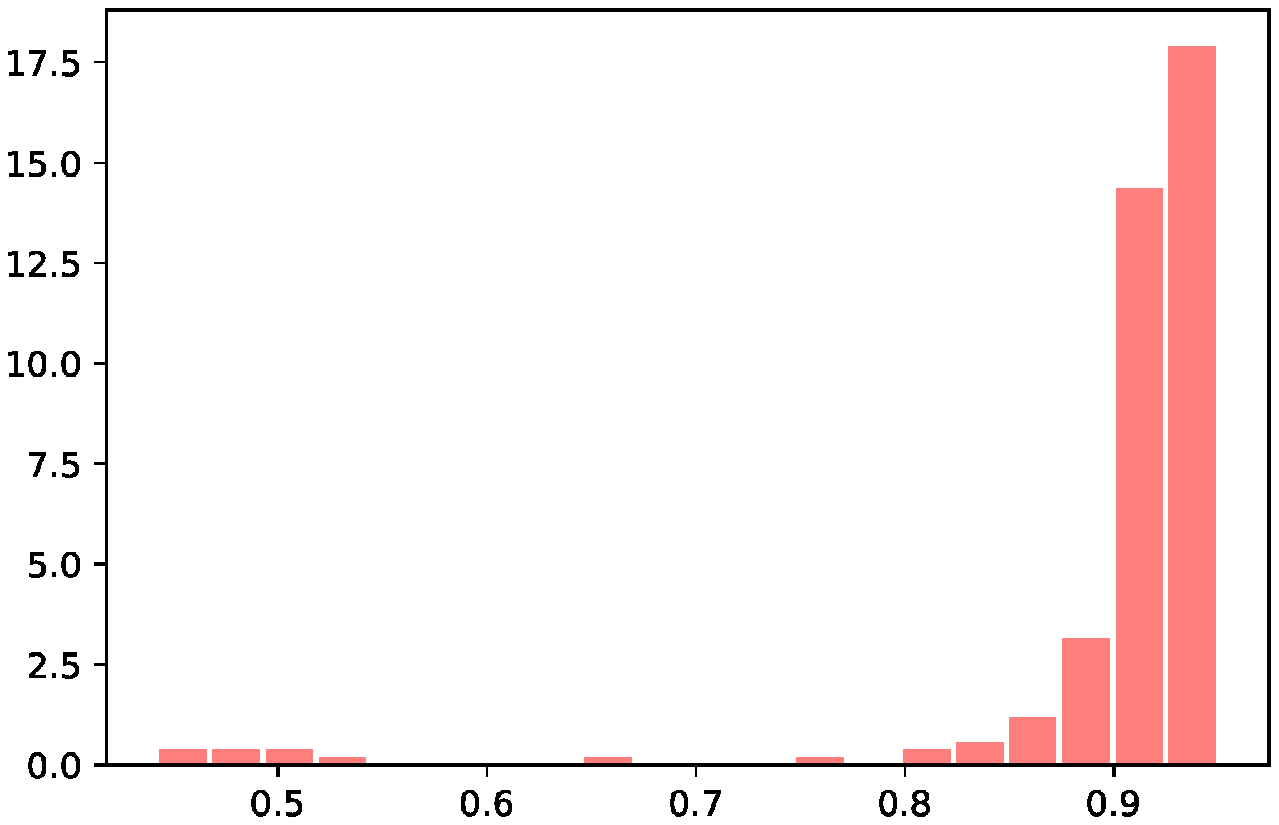}\label{fig_dataset_distribution_cifar10}}
	\hfil
	\subfloat[CIFAR100-CNN]{\includegraphics[width=0.9\columnwidth]{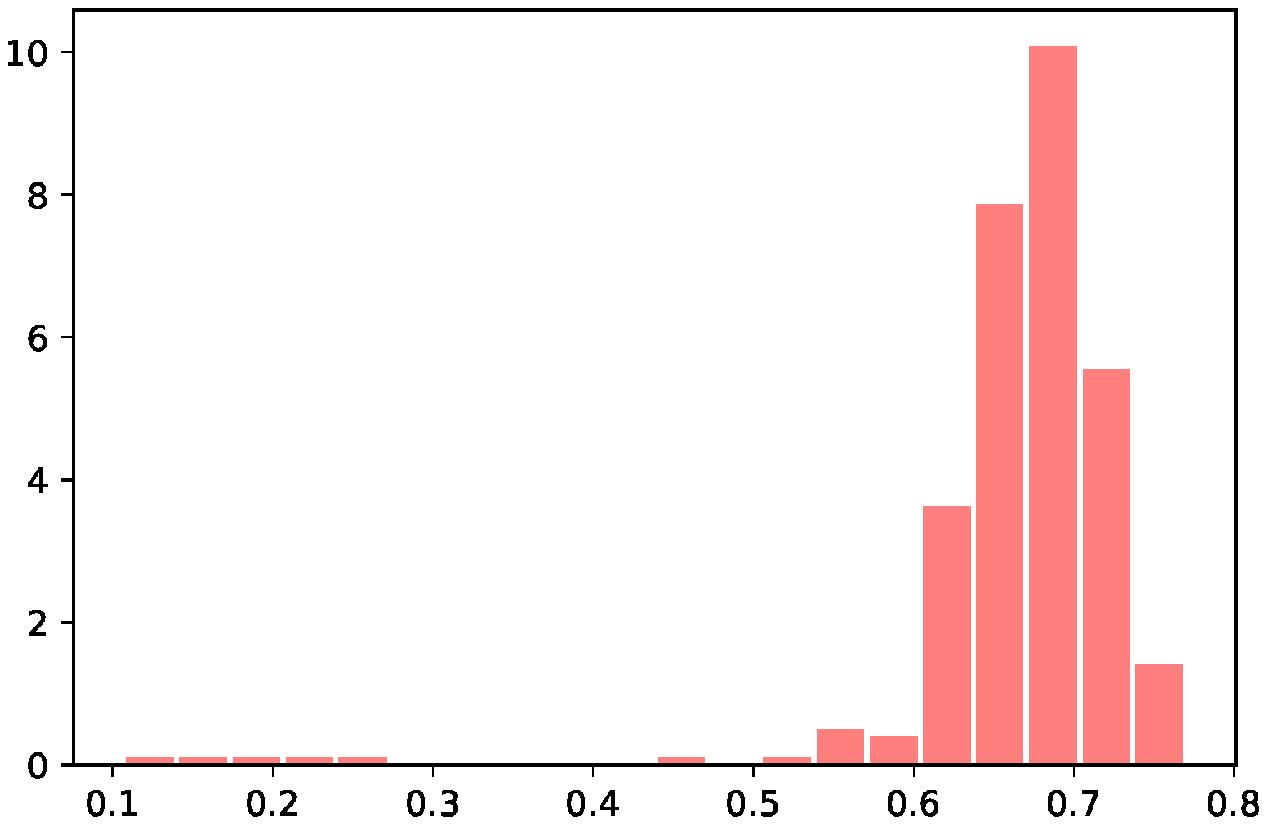}\label{fig_dataset_distribution_cifar100}}
	\caption{The performance distribution of the used datasets with the interval of 0.001, where the horizontal axis denotes the classification accuracy and the vertical axis means the number of architectures. Specifically, Figs.~\ref{fig_dataset_distribution_cifar10} and~\ref{fig_dataset_distribution_cifar100} show the performance distributions of the CNN architectures on CIFAR10 and CIFAR100, respectively (i.e., CIFAR10-CNN and CIFAR100-CNN).}\label{fig_dataset_distribution}
\end{figure*}

Specifically, the datasets are composed of two types of CNN architectures. They are the CNN architectures generated by the AE-CNN algorithm on CIFAR10 and CIFAR100 benchmark datasets~\cite{krizhevsky2009learning}, which are two widely used image classification benchmark datasets for ENAS algorithms. For the convenience of the description, the CNN architectures of CIFAR10 and CIFAR100 are named as CIFAR10-CNN and CIFAR100-CNN, respectively. Particularly, each sample of the CIFAR10-CNN and the CIFAR100-CNN datasets has been trained for 350 epochs on the corresponding dataset with a scheduled learning rate, and the classification accuracy on the validation dataset is used as its performance value. In the datasets, some CNN architectures have the classification accuracy of zeros. This is caused by the out-of-memory errors due to the oversized generated CNN architectures which cannot run on a single GPU card. In this experiment, we only keep the CNN architectures having the classification accuracies greater than zeros. After that, there are total $200$ samples in CIFAR10-CNN, and 299 samples in CIFAR100-CNN. Fig.~\ref{fig_dataset_distribution} show the accuracy distribution of the data, using the interval of the distribution as $0.001$. As can be seen from the figures, most of the samples are with the classification accuracy about $95\%$ for the CIFAR10-CNN dataset, and for the CIFAR100-CNN, while the majority samples are with the classification accuracy of about $70\%$. These values are approximately equal to the best results of NAS algorithms on both datasets, e.g., the AE-CNN algorithm~\cite{sun2019completely}.

\subsection{Regression Models}
\label{sec4_2}

Because the training protocol works based on the performance predictors, which are often realized by regression models, a regression model must be chosen before performing the experiments on the proposed training protocol. In principle, any regression model can be adopted for this purpose. In our experiments, the Support Vector Machine (SVM)~\cite{suykens1999least}, the Gradient Boosting Decision Tree (GBDT)~\cite{friedman2001greedy}, the Decision Tree (DTree)~\cite{quinlan1996learning}, and the Random Forest (RForest)~\cite{liaw2002classification} are chosen as the regression models.

Firstly, instead of adopting only one of them, we choose the four regression models to collectively accomplish the experiments. It is expected that the experimental results of the proposed training protocol will not be biased by the specific regression model adopted. Secondly, the four chosen regression models are very popular among the machine learning community, and most researchers are familiar with them. Using them as the regression models to conduct the experiments, the experimental results should be better understandable, and the effectiveness of the proposed training protocol will also be easier justified. Thirdly, all the four chosen regression models have the corresponding versions to accomplish the experiments of the baseline and the proposed training protocol. Specifically, the baseline, i.e., the traditional training protocol, works on the linear regression model, while the proposed training protocol works on the logistic regression model. The four chosen regression models have the corresponding linear regression versions and also the logistic regression versions. Please note that the neural network model is not chosen in the experiments, although it has also been popular among the machine learning community. The main reason is that the neural network model has a large number of hyperparameters to be tuned for guaranteeing its promising performance, while comprehensive investigating the combinations of the hyperparameters is impractical and prohibitively costly. This is beyond the scope of this study. Please note that we use the implementations  provided by the Sklearn library~\cite{pedregosa2011scikit} for the four chosen models.

\subsection{Parameter Settings}
\label{sec4_3}
In this subsection, we will provide the details about the parameter settings for the experiments, including the settings for constructing the training data and the test data from the corresponding dataset, and the settings for the parameters of the chosen regression models.

\subsubsection{Construction of Training Data  and Test Data}

Commonly, the dataset is randomly shuffled, and then the training data and the test data are split with a specified proportion based on the conventions of the machine learning community. In this experiments, we choose the proportion of $7:3$ to split the dataset, i.e., the training data and the test data account for the $70\%$ and $30\%$ of the corresponding dataset, respectively. Commonly, the dataset is random shuffled and the splitting is performed. In our experiments, the training data and the test data are splitted based on the real scenario of the ENAS algorithms, i.e., the details which have been discussed in Subsection~\ref{sec2_2}. To achieve this, we first sort the samples from the dataset based on their performance values in an ascending order, and then make the first $70\%$ samples of the dataset as the training data, while the remains are left to the test data. As have discussed above, we employ the vectorization method introduced in E2EPP to realize the dataset into the proper format, which can be directly used by the regression models. As a result, each CNN, after the vectorization, is composed of 32 features represented by real numbers. Among these numbers, the first 31 features are integers denoting the corresponding architecture information, and the last one is a floating point number referring to its performance (a number between 0 and 1). The training data and the test data of the chosen dataset can be downloaded from~\url{ https://github.com/yn-sun/PRI}. Please note that we did not construct the validation data in the experiments. That is because we will use the cross-validation strategy~\cite{schaffer1993selecting} to choose the best hyperparameters for the chosen regression models, and the corresponding validation operation will be performed on the data split from the training data. More details can be found in the next subsection.

 \begin{table*}
 	\renewcommand{\arraystretch}{1.2}
 	\caption{The comparisons between the proposed training protocol and the baseline with the four chosen regression models. The results are the classification accuracies on the CIFAR10-CNN and CIFAR100-CNN benchmark datasets with the corresponding regression models}
 	\label{tab_overview}
 	\center
 	\begin{tabular}{c|c|c|c|c}
 		\hline
 		&\multicolumn{2}{c|}{\textbf{CIFAR10-CNN}}&\multicolumn{2}{c}{\textbf{CIFAR100-CNN}}\\
 		
 		\hline
 		& Baseline & The Proposed Training Protocol & Baseline & The Proposed Training Protocol\\
 		\hline
 		\textbf{SVM}&47.91\%&65.59\%&39.68\%&64.81\%\\
 		\hline
 		\textbf{GBDT}&55.48\%&60.65\%&51.94\%&58.16\%\\
 		\hline
 		\textbf{DTree}&40.79\%&54.77\%&49.69\%&54.96\%\\
 		\hline
 		\textbf{RForest}&50.11\%&63.25\%&52.46\%&55.39\%\\
 		
 		\hline
 	\end{tabular}
 \end{table*}

\subsubsection{Parameter Settings for the Chosen Regression Models}
The performance of almost all machine learning algorithms is sensitive to their parameter settings. However, these parameter settings are often problem-specified. To obtain a fair comparison in our experiments, we will perform the search for choosing the best parameter settings of each regression model employed before running the corresponding model on the chosen dataset. Specifically, we will first specify a range for each parameter of the four regression models based on the conventions, and then use the random search strategy to try each combination of the parameters. The model with the corresponding parameter combination will be trained and evaluated via the 5-fold cross validation. Finally, the best parameter combination is chosen for performing the comparison between the proposed training protocol and the baseline. In the following, the search ranges of the candidate parameters are provided.

The four chosen regression models have multiple parameters affecting their performance. However, in practice, only some of them need to be extensively tuned, and others are just kept as their default values defined in the used library. Specifically, The Dtree model has the following parameters needed to be tuned: 1) the minimum number of samples required to split an internal node; 2) the minimum number of samples required to be a leaf node; 3) the maximum depth of the tree; 4) the number of features to consider for the best split; 5) the criterion to evaluate the model. For the convenience of the description, these five parameters are denoted as $min\_samples\_split$, $min\_samples\_leaf$, $max\_depth$, $max\_features$, and $criterion$, respectively. Based on the conventions of the community, the search ranges of $min\_samples\_split$ and $max\_depth$ are both set to $[2..200]$, the search ranges of $min\_samples_leaf$ and $n\_estimators$ are set to $[1..300]$ and $[1..200]$, respectively, and the $max\_features$ is chosen from $\{\sqrt{n\_features}$, $log2(n\_features)\}$ where $n\_featuees$ refers to the number of the features. The $crtiterion$ is set based on the situation where the corresponding model is used. Particularly, when the model is used for the proposed training protocol, the value of $criterion$ is chosen from the Gini impurity~\cite{simovici2002impurity} and the information gain~\cite{kent1983information}. While when the mode is used for the baseline, the $criterion$ is chosen from the mean squared error and the mean absolute error. Please note that the choice of ``$criterion$'' is based on the principle design of both types of regression models.

The GBDT is also an ensemble method based on the decision trees~\cite{Quinlan} as the RForest does. Their intuitive difference remains on the way as how to organize the decision trees. In the RForest model, the trees are constructed in parallel, In contrast, the trees in the GBDT model are sequentially constructed through the boosting~\cite{mason2000boosting}, where the subsequent one focuses on complementing the performance loss resulted by the previous one. In addition, the trees in the GBDT are organized through the optimization of the gradient-based method. As a result, the parameters tuned for the RForest are also those of the GBDT, and their search ranges are also specified in the same manner. In addition, the GBDT also has the parameter of the learning rate regarding the corresponding gradient-based optimization, and the search range is specified from $1,000$ random variables sampled from the exponential distribution scaled within $10$. As for the SVM, the parameters to be tuned are the kernel type, the kernel coefficient, and the regularization parameter. In our experiment, the radial basis function~\cite{musavi1992training} is used as the kernel given its popularity among the community and the effectiveness in addressing real-world problems. For the other two parameters, their search ranges are set as the same as the learning rate of the GBDT.

 \begin{table*}
	\renewcommand{\arraystretch}{1.2}
	\caption{The ablation experiment results of the proposed components of the proposed training protocol.}
	\label{tab_ablition}
	\center
	\begin{tabular}{c|c|c|c|c|c|c}
		\hline
		&\multicolumn{3}{c|}{\textbf{CIFAR10-CNN}}&\multicolumn{3}{c}{\textbf{CIFAR100-CNN}}\\
		
		\hline
		& The Proposed Training Protocol& G1 & G2 & The Proposed Training Protocol & G1 & G2\\
		\hline
		\textbf{SVM}&65.59\%&48.02\%&63.28\%&64.81\%&53.73\%&64.49\%\\
		\hline
		\textbf{GBDT}&60.65\%&43.33\%&58.38\%&58.16\%&51.66\%&55.61\%\\
		\hline
		\textbf{DTree}&54.77\%&43.67\%&49.21\%&54.96\%&50.21\%&53.33\%\\
		\hline
		\textbf{RForest}&63.25\%&41.36\%&62.40\%&55.39\%&47.77\%&55.19\%\\
		\hline
		\hline
		\textbf{Avg}&61.07\%&44.10\%&58.32\%&58.33\%&50.84\%&57.16\%\\
		\hline
	\end{tabular}
\end{table*}

\subsection{Performance Metrics}
\label{sec4_4}

As have justified in Subsection~\ref{sec2_2}, the goal of the performance predictors is to compare the ordering of the candidates, to guide the search of the corresponding EC method. In the proposed training protocol, this is achieved by the proposed PRI to compare the ordering of two candidates. The reason for specifying two is that when the comparison is required to operate on the candidacies with the size greater than two, the PRI can still achieve that by just performing it multiple times. The PRI is consistent with the mechanism of the performance predictors in addressing the real situations, and can be formalized as a logistic regression problem, i.e., the binary classification problem. As a result, we choose the classification accuracy as the metric in the experiments, i.e., to check how many the trained performance predictors can successfully predict the ground-truth ordering between any two test data. The higher the classification accuracy, the better the corresponding training protocol.

\section{Experiment Results}
\label{sec5}
\subsection{Overall Results}

The experimental results of the proposed training protocol against those of the baseline are shown in Table~\ref{tab_overview}. Specifically, in Table~\ref{tab_overview}, the first column denotes the names of the chosen regression models employed by the baseline and the proposed training protocol. The second and the fourth columns show the classification accuracies achieved by the baseline on the CIFAR10-CNN and the CIFAR100-CNN benchmark datasets, respectively, while the third and the fifth columns list the classification accuracies of the proposed training protocol with the four regression models on the CIFAR10-CNN and the CIFAR100-CNN benchmark datasets, respectively.

As can be seen from Table~\ref{tab_overview}, when the SVM is employed as the regression model, the proposed training protocol achieves the classification accuracy of 65.59\%. This is superior to that of the baseline model, which is with the classification accuracy of 47.91\%. This gap becomes even more significant on the CIFAR100-CNN benchmark dataset, where the proposed training protocol obtains the classification accuracy of 64.81\%, while the baseline only achieves the classification accuracy of 39.68\%. When the GBDT is utilized as the regression model, the proposed training protocol receives the classification accuracy of 60.65\% and 58.16\% on CIFAR10-CNN and CIFAR100-CNN, respectively, while the baseline merely achieves those of 55.48\% and 51.94\%, respectively. In addition, the baseline obtains the classification accuracies of 40.79\% and 49.69\%, with the DTree regression model on CIFAR10-CNN and CIFAR100-CNN, respectively. Clearly, both are inferior to the proposed training protocol which achieves the classification accuracies of 54.77\% on CIFAR10-CNN and 54.96\% on CIFAR100-CNN. With the RForest regression model, the baseline obtains the classification accuracy of 50.11\% on CIFAR10-CNN, and 52.46\% on CIFAR100-CNN, while the proposed training protocol achieves the classification accuracies of 63.25\% and 55.39\% on CIFAR10-CNN and CIFAR100-CNN, respectively. To summarize, the proposed training protocol outperforms the baseline with the four popular regression models on the two chosen benchmark datasets.

\subsection{Ablation Experiments}
In order to further check how much the proposed three components affect the overall performance of the proposed training protocol, the ablation experiments are reported in this subsection, and the experimental results are discussed. As have mentioned, there are three critical components in the proposed training protocol, i.e., 1) the PRI using the order of any two samples indicating the target to training the samples, 2) the utilization of the logistic regression that promotes an accurate regression model to be built and 3) the use of the difference between any two samples as the training instances, for constructing a balanced training dataset for the performance predictors. Because the first two components work inter-dependently and cannot function independently if any one disabled. As a result, in this experiment, we put both together as a component for the ablation experiments.

Specifically, we have performed two groups of the experiments for the goal of the ablation. The first is achieved by disabling the first two components from the proposed training protocol (denoted as ``G1''), and the second is conducted by randomly choosing only one training sample from the positive or negative samples (denoted as ``G2'').  Table~\ref{tab_ablition} shows the results of the ablation experiments of the proposed training protocol on the CIFAR10-CNN and the CIFAR100-CNN datasets. As can be seen from Table~\ref{tab_ablition}, the third and sixth columns provide the experimental results in terms of disabling the first two components of the proposed training protocol, while the fourth and the seventh columns show those of disabling the third component of the proposed training protocol. In addition, the third rows to the sixth rows of Table~\ref{tab_ablition} list the corresponding experimental results on the four chosen regression models. For the convenience of summarizing the performance contribution of the corresponding component(s), the last row provides the average results of the four regression models on the same benchmark dataset.

As can be seen from Table~\ref{tab_ablition}, for the CIFAR10-CNN benchmark, when the first two components are disabled from the proposed training protocol, the SVM, the GBDT, the DTree and the RForest obtain the results of 48.02\%, 43.33\%, 43,67\%, and 41.36\%, respectively. While only not using the third component, all the results show better accuracies, i.e., the results of 63.28\%, 58.38\%, 49.21\%, and 62.40\% are achieved from the respective regression model. This comparison result can also be observed from the CIFAR100-CNN benchmark dataset. Particularly, when the first two components are disabled from the proposed training protocol, the SVM, the GBDT, the DTree, and the RForest achieve the results of 53.73\%, 51.66\%, 50.21\%, and 47.77\%, respectively. When the third component is disabled from the proposed training protocol, these four regression models can achieve the results of 64.49\%, 55.61\%, 53.33\% and 55.19, respectively. The conclusion can be easily drawn from the number listed in the last row of Table~\ref{tab_ablition}. Specifically, if the first two components are not used by the proposed training protocol, the performance of the proposed training protocol will be deteriorated by 16.97\% on the CIFAR10-CNN benchmark dataset and 7.49\% on the CIFAR100-CNN benchmark dataset. However, when only the third component is disabled from the proposed training protocol, the deterioration is only 2.75\% and 1.17\%, respectively. Clearly, the first two components contributed more to the performance of the proposed training protocol.

\section{Conclusion and Future Work}
\label{sec6}
The goal of this paper is to propose an effective and efficient training protocol for performance predictors of evolutionary neural architecture search algorithms. The goal has been achieved by the three proposed components. Firstly, we have proposed a performance ranking indicator to be as the training target. The indicator is consistent with the real scenario when the performance predictors are used for real-world applications, thus being able to provide satisfactory prediction results. Secondly, we have proposed using the logistic regression to replace the linear regression in the traditional training protocol, which is able to alleviate the difficulty in designing an exact regression model to predict the order of any two samples. Thirdly, we have proposed to use the difference between any two samples as the training instance, which is naturally without the data imbalanced problem that inadvertently challenges the accuracy of the corresponding regression model. The proposed training protocol is investigated by four popular regression models against the traditional training protocol on two datasets. The experimental results have demonstrated the effectiveness of the proposed training protocol. In addition, we have also conducted the ablation experiments and found that the proposed first two components contribute more to the performance of the proposed training protocol than the third dose. In future, we will extend efforts in designing a unified vectorization method for neural network architectures, with the expectation that the proposed training protocol can be easily adopted by any evolutionary neural architecture search algorithms.



\begin{thebibliography}{10}
	\providecommand{\url}[1]{#1}
	\csname url@samestyle\endcsname
	\providecommand{\newblock}{\relax}
	\providecommand{\bibinfo}[2]{#2}
	\providecommand{\BIBentrySTDinterwordspacing}{\spaceskip=0pt\relax}
	\providecommand{\BIBentryALTinterwordstretchfactor}{4}
	\providecommand{\BIBentryALTinterwordspacing}{\spaceskip=\fontdimen2\font plus
		\BIBentryALTinterwordstretchfactor\fontdimen3\font minus
		\fontdimen4\font\relax}
	\providecommand{\BIBforeignlanguage}[2]{{%
			\expandafter\ifx\csname l@#1\endcsname\relax
			\typeout{** WARNING: IEEEtran.bst: No hyphenation pattern has been}%
			\typeout{** loaded for the language `#1'. Using the pattern for}%
			\typeout{** the default language instead.}%
			\else
			\language=\csname l@#1\endcsname
			\fi
			#2}}
	\providecommand{\BIBdecl}{\relax}
	\BIBdecl
	
	\bibitem{lecun2015deep}
	Y.~LeCun, Y.~Bengio, and G.~Hinton, ``Deep learning,'' \emph{nature}, vol. 521,
	no. 7553, pp. 436--444, 2015.
	
	\bibitem{hinton2006reducing}
	G.~E. Hinton and R.~R. Salakhutdinov, ``Reducing the dimensionality of data
	with neural networks,'' \emph{science}, vol. 313, no. 5786, pp. 504--507,
	2006.
	
	\bibitem{krizhevsky2012imagenet}
	A.~Krizhevsky, I.~Sutskever, and G.~E. Hinton, ``Imagenet classification with
	deep convolutional neural networks,'' in \emph{Advances in neural information
		processing systems}, 2012, pp. 1097--1105.
	
	\bibitem{kearney1987optical}
	J.~K. Kearney, W.~B. Thompson, and D.~L. Boley, ``Optical flow estimation: An
	error analysis of gradient-based methods with local optimization,''
	\emph{IEEE Transactions on Pattern Analysis and Machine Intelligence}, no.~2,
	pp. 229--244, 1987.
	
	\bibitem{he2016deep}
	K.~He, X.~Zhang, S.~Ren, and J.~Sun, ``Deep residual learning for image
	recognition,'' in \emph{Proceedings of 2016 IEEE Conference on Computer
		Vision and Pattern Recognition}, Las Vegas, NV, USA, 2016, pp. 770--778.
	
	\bibitem{huang2017densely}
	G.~Huang, Z.~Liu, K.~Q. Weinberger, and L.~van~der Maaten, ``Densely connected
	convolutional networks,'' in \emph{Proceedings of 2017 IEEE Conference on
		Computer Vision and Pattern Recognition}, Honolulu, HI, USA, 2017, pp.
	2261--2269.
	
	\bibitem{elsken2018neural}
	T.~Elsken, J.~H. Metzen, and F.~Hutter, ``Neural architecture search: A
	survey,'' \emph{arXiv preprint arXiv:1808.05377}, 2018.
	
	\bibitem{sun2019completely}
	Y.~Sun, B.~Xue, M.~Zhang, and G.~G. Yen, ``Completely automated cnn
	architecture design based on blocks,'' \emph{IEEE transactions on neural
		networks and learning systems}, vol.~31, no.~4, pp. 1242--1254, 2020.
	
	\bibitem{baker2016designing}
	B.~Baker, O.~Gupta, N.~Naik, and R.~Raskar, ``Designing neural network
	architectures using reinforcement learning,'' in \emph{Proceedings of the
		2017 International Conference on Learning Representations}, Toulon, France,
	2017.
	
	\bibitem{zoph2016neural}
	B.~Zoph and Q.~V. Le, ``Neural architecture search with reinforcement
	learning,'' in \emph{Proceedings of the 2017 International Conference on
		Learning Representations}, Toulon, France, 2017.
	
	\bibitem{liu2018darts}
	H.~Liu, K.~Simonyan, and Y.~Yang, ``Darts: Differentiable architecture
	search,'' \emph{arXiv preprint arXiv:1806.09055}, 2018.
	
	\bibitem{xie2017genetic}
	L.~Xie and A.~Yuille, ``Genetic {CNN},'' in \emph{Proceedings of 2017 IEEE
		International Conference on Computer Vision}, Venice, Italy, 2017, pp.
	1388--1397.
	
	\bibitem{real2017large}
	E.~Real, S.~Moore, A.~Selle, S.~Saxena, Y.~L. Suematsu, J.~Tan, Q.~Le, and
	A.~Kurakin, ``Large-scale evolution of image classifiers,'' in
	\emph{Proceedings of 2017 Machine Learning Research}, Sydney, Australia,
	2017, pp. 2902--2911.
	
	\bibitem{liu2017hierarchical}
	H.~Liu, K.~Simonyan, O.~Vinyals, C.~Fernando, and K.~Kavukcuoglu,
	``Hierarchical representations for efficient architecture search,'' in
	\emph{Proceedings of 2018 Machine Learning Research}, Stockholm, Sweden,
	2018.
	
	\bibitem{suganuma2017genetic}
	M.~Suganuma, S.~Shirakawa, and T.~Nagao, ``A genetic programming approach to
	designing convolutional neural network architectures,'' in \emph{Proceedings
		of the 2017 Genetic and Evolutionary Computation Conference}.\hskip 1em plus
	0.5em minus 0.4em\relax Berlin, Germany: ACM, 2017, pp. 497--504.
	
	\bibitem{sun2019evolving}
	Y.~Sun, B.~Xue, M.~Zhang, and G.~G. Yen, ``Evolving deep convolutional neural
	networks for image classification,'' \emph{IEEE Transactions on Evolutionary
		Computation}, vol.~24, no.~2, pp. 394--407, 2020.
	
	\bibitem{back1996evolutionary}
	T.~Back, \emph{Evolutionary Algorithms in Theory and Practice: Evolution
		Strategies, Evolutionary Programming, Genetic Algorithms}.\hskip 1em plus
	0.5em minus 0.4em\relax England, UK: Oxford university press, 1996.
	
	\bibitem{banzhaf1998genetic}
	W.~Banzhaf, P.~Nordin, R.~E. Keller, and F.~D. Francone, \emph{Genetic
		Programming: An Introduction}.\hskip 1em plus 0.5em minus 0.4em\relax Morgan
	Kaufmann San Francisco, 1998.
	
	\bibitem{schmitt2001theory}
	L.~M. Schmitt, ``Theory of genetic algorithms,'' \emph{Theoretical Computer
		Science}, vol. 259, no. 1-2, pp. 1--61, 2001.
	
	\bibitem{wang2015two_arch2}
	H.~Wang, L.~Jiao, and X.~Yao, ``Two\_arch2: an improved two-archive algorithm
	for many-objective optimization,'' \emph{IEEE Transactions on Evolutionary
		Computation}, vol.~19, no.~4, pp. 524--541, 2015.
	
	\bibitem{sun2018igd}
	Y.~Sun, G.~G. Yen, and Z.~Yi, ``{IGD} indicator-based evolutionary algorithm
	for many-objective optimization problems,'' \emph{IEEE Transactions on
		Evolutionary Computation}, vol.~23, no.~2, pp. 173--187, 2019.
	
	\bibitem{twostage2019}
	Y.~Sun, B.~Xue, M.~Zhang, and G.~G. Yen, ``A new two-stage evolutionary
	algorithm for many-objective optimization,'' \emph{IEEE Transactions on
		Evolutionary Computation}, vol.~23, no.~5, pp. 748--761, 2019.
	
	\bibitem{krizhevsky2009learning}
	A.~Krizhevsky, G.~Hinton \emph{et~al.}, ``Learning multiple layers of features
	from tiny images,'' 2009.
	
	\bibitem{wang2018evolving}
	B.~Wang, Y.~Sun, B.~Xue, and M.~Zhang, ``Evolving deep convolutional neural
	networks by variable-length particle swarm optimization for image
	classification,'' in \emph{2018 IEEE Congress on Evolutionary Computation
		(CEC)}.\hskip 1em plus 0.5em minus 0.4em\relax IEEE, 2018, pp. 1--8.
	
	\bibitem{swersky2014freeze}
	K.~Swersky, J.~Snoek, and R.~P. Adams, ``Freeze-thaw bayesian optimization,''
	\emph{arXiv preprint arXiv:1406.3896}, 2014.
	
	\bibitem{domhan2015speeding}
	T.~Domhan, J.~T. Springenberg, and F.~Hutter, ``Speeding up automatic
	hyperparameter optimization of deep neural networks by extrapolation of
	learning curves.'' in \emph{Proceedings of the 24th International Conference
		on Artificial Intelligence}, vol.~15, 2015, pp. 3460--8.
	
	\bibitem{klein2016learning}
	A.~Klein, S.~Falkner, J.~T. Springenberg, and F.~Hutter, ``Learning curve
	prediction with {B}ayesian neural networks,'' \emph{Proceedings of the 5th
		International Conference on Learning Representations}, 2016.
	
	\bibitem{pham2018efficient}
	H.~Pham, M.~Y. Guan, B.~Zoph, Q.~V. Le, and J.~Dean, ``Efficient neural
	architecture search via parameter sharing,'' \emph{Proceedings of the 35th
		International Conference on Machine Learning}, vol.~80, pp. 4095--4104, 2018.
	
	\bibitem{istrate2018tapas}
	R.~Istrate, F.~Scheidegger, G.~Mariani, D.~Nikolopoulos, C.~Bekas, and A.~C.~I.
	Malossi, ``{TAPAS}: train-less accuracy predictor for architecture search,''
	\emph{arXiv preprint arXiv:1806.00250}, 2018.
	
	\bibitem{baker2017accelerating}
	B.~Baker, O.~Gupta, R.~Raskar, and N.~Naik, ``Accelerating neural architecture
	search using performance prediction,'' \emph{2018 International Conference on
		Learning Representations Workshop}, 2017.
	
	\bibitem{deng2017peephole}
	B.~Deng, J.~Yan, and D.~Lin, ``Peephole: Predicting network performance before
	training,'' \emph{arXiv preprint arXiv:1712.03351}, 2017.
	
	\bibitem{sun2019surrogate}
	Y.~Sun, H.~Wang, B.~Xue, Y.~Jin, G.~G. Yen, and M.~Zhang, ``Surrogate-assisted
	evolutionary deep learning using an end-to-end random forest-based
	performance predictor,'' \emph{IEEE Transactions on Evolutionary
		Computation}, vol.~24, no.~2, pp. 350--364, 2020.
	
	\bibitem{de1975analysis}
	K.~A. De~Jong, ``Analysis of the behavior of a class of genetic adaptive
	systems,'' Tech. Rep., 1975.
	
	\bibitem{miller1995genetic}
	B.~L. Miller, D.~E. Goldberg \emph{et~al.}, ``Genetic algorithms, tournament
	selection, and the effects of noise,'' \emph{Complex systems}, vol.~9, no.~3,
	pp. 193--212, 1995.
	
	\bibitem{abadi2016tensorflow}
	M.~Abadi, P.~Barham \emph{et~al.}, ``Tensorflow: A system for large-scale
	machine learning,'' in \emph{12th $\{$USENIX$\}$ Symposium on Operating
		Systems Design and Implementation ($\{$OSDI$\}$ 16)}, 2016, pp. 265--283.
	
	\bibitem{paszke2019pytorch}
	A.~Paszke, S.~Gross \emph{et~al.}, ``Pytorch: An imperative style,
	high-performance deep learning library,'' in \emph{Advances in Neural
		Information Processing Systems}, 2019, pp. 8024--8035.
	
	\bibitem{suykens1999least}
	J.~A. Suykens and J.~Vandewalle, ``Least squares support vector machine
	classifiers,'' \emph{Neural processing letters}, vol.~9, no.~3, pp. 293--300,
	1999.
	
	\bibitem{friedman2001greedy}
	J.~H. Friedman, ``Greedy function approximation: a gradient boosting machine,''
	\emph{Annals of statistics}, pp. 1189--1232, 2001.
	
	\bibitem{quinlan1996learning}
	J.~R. Quinlan, ``Learning decision tree classifiers,'' \emph{ACM Computing
		Surveys (CSUR)}, vol.~28, no.~1, pp. 71--72, 1996.
	
	\bibitem{liaw2002classification}
	A.~Liaw, M.~Wiener \emph{et~al.}, ``Classification and regression by
	randomforest,'' \emph{R news}, vol.~2, no.~3, pp. 18--22, 2002.
	
	\bibitem{pedregosa2011scikit}
	F.~Pedregosa, G.~Varoquaux, A.~Gramfort, V.~Michel, B.~Thirion, O.~Grisel,
	M.~Blondel, P.~Prettenhofer, R.~Weiss, V.~Dubourg \emph{et~al.},
	``Scikit-learn: Machine learning in python,'' \emph{the Journal of machine
		Learning research}, vol.~12, pp. 2825--2830, 2011.
	
	\bibitem{schaffer1993selecting}
	C.~Schaffer, ``Selecting a classification method by cross-validation,''
	\emph{Machine Learning}, vol.~13, no.~1, pp. 135--143, 1993.
	
	\bibitem{simovici2002impurity}
	D.~A. Simovici, D.~Cristofor, and L.~Cristofor, ``Impurity measures in
	databases,'' \emph{Acta Informatica}, vol.~38, no.~5, pp. 307--324, 2002.
	
	\bibitem{kent1983information}
	J.~T. Kent, ``Information gain and a general measure of correlation,''
	\emph{Biometrika}, vol.~70, no.~1, pp. 163--173, 1983.
	
	\bibitem{Quinlan}
	J.~R. Quinlan, ``Induction of decision trees,'' \emph{Machine learning},
	vol.~1, no.~1, pp. 81--106, 1986.
	
	\bibitem{mason2000boosting}
	L.~Mason, J.~Baxter, P.~L. Bartlett, and M.~R. Frean, ``Boosting algorithms as
	gradient descent,'' in \emph{Advances in neural information processing
		systems}, 2000, pp. 512--518.
	
	\bibitem{musavi1992training}
	M.~T. Musavi, W.~Ahmed, K.~H. Chan, K.~B. Faris, and D.~M. Hummels, ``On the
	training of radial basis function classifiers,'' \emph{Neural networks},
	vol.~5, no.~4, pp. 595--603, 1992.
	
\end{thebibliography}
\end{document}